\documentclass{article}
\usepackage{arxiv}

\usepackage[utf8]{inputenc} 
\usepackage[T1]{fontenc}    
\usepackage{hyperref}       
\usepackage{url}            
\usepackage{booktabs}       
\usepackage{amsfonts}       
\usepackage{nicefrac}       
\usepackage{microtype}      
\usepackage{lipsum}
\usepackage{graphicx}
\graphicspath{ {./images/} }
\usepackage{amssymb}
\usepackage{comment}
\usepackage{amsmath}

\usepackage{multirow}
\usepackage{siunitx}
\usepackage{threeparttable}

\usepackage{geometry}
\usepackage{subcaption}
\usepackage{usbib}
\title{The influence of missing data mechanisms and simple missing data handling techniques on fairness}

\author{
 Aeysha Bhatti \\
  Department of Statistics and Actuarial Science\\
  Stellenbosch University\\
  Stellenbosch, 7602 \\
  South Africa\\
  \texttt{22456910@sun.ac.za} \\
   \And
  Trudie Sandrock \\
  Department of Statistics and Actuarial Science\\
  Stellenbosch University\\
  Stellenbosch, 7602 \\
  South Africa\\
  \texttt{trudies@sun.ac.za} \\
  \And
  Johan\'e Nienkemper-Swanepoel \\
  Department of Statistics and Actuarial Science\\
  Stellenbosch University\\
  Stellenbosch, 7602 \\
  South Africa\\
  \texttt{nienkemperj@sun.ac.za} \\
}

\begin{document}
\maketitle
\begin{abstract}
Machine learning algorithms permeate the day-to-day aspects of our lives and therefore studying the fairness of these algorithms before implementation is crucial. One way in which bias can manifest in a dataset is through missing values. Missing data are often assumed to be missing completely randomly; in reality the propensity of data being missing is often tied to the demographic characteristics of individuals. There is limited research into how missing values and the handling thereof can impact the fairness of an algorithm. Most researchers either apply listwise deletion or tend to use simpler methods of imputation (e.g. mean or mode) compared to more advanced approaches (e.g. multiple imputation). This study considers the fairness of various classification algorithms after a range of missing data handling strategies is applied. Missing values are generated (i.e. amputed) in three popular datasets for classification fairness, by creating a high percentage of missing values using three missing data mechanisms. The results show that the missing data mechanism does not significantly impact fairness; across the missing data handling techniques listwise deletion gives the highest fairness on average and amongst the classification algorithms random forests leads to the highest fairness on average. The interaction effect of the missing data handling technique and the classification algorithm is also often significant.
\end{abstract}


\section{INTRODUCTION}
The pervasiveness of machine learning (ML) algorithms in our everyday lives is undeniable. The breadth of examples is substantial and includes algorithms used to award or deny bank loans, manage admission into educational institutions, considerably speed up the process of hiring and even aid law enforcement to decide the severity of sentencing of criminals \citep{joy, zlio, book}. Some of the more seemingly innocuous examples include algorithms making movie recommendations, dating websites recommending a suitable partner choice for users and retail websites providing shopping recommendations \citep{mehrabi, hardt, 2020bias}. 
\\
\\
There are many worthwhile reasons to transfer decision-making from humans to computers: they are faster, they can deal with much larger amounts of information more efficiently and accurately, they do not get bored and overall, they reduce human error \citep{book, 2020bias, mehrabi}. Despite these advantages, there are many instances where these algorithms have acted far from justly, with respect to certain sensitive or protected attributes such as race, sex or age. An example is the Correctional Offender Management Profiling for Alternative Sanctions (COMPAS) system, which is used by courts in the United States to assess the risk of re-offence, which gave higher risk values for black offenders and lower risk values for white offenders than their actual risk. Further examples include facial recognition software which has the lowest accuracy on females who are darker skinned, and the online advertisement platform Google Ads which showed considerably fewer advertisements for high-paying jobs to women than to men \citep{joy, 2020bias, mehrabi}.
\\
\\
One of the main reasons for the biased behaviour of algorithms is not only the quality of data that they are trained on, such that historic human biases for or against certain groups are inherent in many datasets; the under representation of minorities in datasets often means data pertaining to them is missing. ML algorithms, which are designed to detect and learn patterns in data, exploit and can even amplify such human biases in making decisions. Reduced data quantity for minorities and hence class imbalance can lead to reduced accuracy for minorities, leading to further discrimination by the algorithms.
\\
\\
An important way in which bias manifests in a dataset is through the absence of certain fields or inputs. Missing data refers to those instances of data which contain fields which have not been captured or have been lost or deleted \citep{enders}. If data are missing, these data are often assumed to be missing completely randomly, but usually this is not the case, and the propensity of data being missing is often tied to socio-economic status or demographic characteristics of individuals \citep{ugduck}. Despite commendable advances in missing data handling in research \citep{ich}, the reasons for missingness are usually not taken into account before proceeding with the treatment of missing data \citep{missingmethodsprop}.
\\
\\
In practice, missing data are often dealt with in one of two ways before any statistical analysis can be performed. The first of these is listwise deletion (LD), which refers to complete removal of the observation unit (i.e. sample) which contains the missing fields. The second way is imputation, where at least one value is presented as a placeholder for the field where there is a missing value. It is plausible that missing data and bias of an algorithm are related; however, limited research on fairness of algorithms in the context of simpler missing data data techniques are available \citep{ugduck, zhanglong, wangsingh, caton}. For example, if LD is implemented and there are many more observations with missing fields for one group over another, the algorithm might give more accurate predictions for the group with more observed information and perform comparatively worse for the other group. When applied in a real world context, these biased decisions could have a tangible and detrimental impact on an individual \citep{joy, wmd, airbnb}.
\\
\\
Research at the intersection of ML fairness and missing values is relatively new, but deserves attention due to the importance of the application of the methodology. Foundational aspects of the research presented in this manuscript are addressed to complement existing work, by focusing on the impact of missing data and handling procedures on the fairness and accuracy of ML algorithms. Important considerations to understand this impact, are the cause of missingness (i.e. the missing data mechanism) and the typical approaches to deal with the missing data (i.e. LD and imputation). Since simpler missing data handling methods are popular and widely used by researchers \citep{vim, missingmethodsprop}, it is important to study and report on the performance of these techniques when applied in the context of the missing data mechanisms and in combination with various classifiers. While multiple imputation (MI) remains a superior approach to handle missing values, currently methodology does not exist to appropriately pool and analyse fairness metric estimates obtained in this way. Therefore, comparing the performance of SI and MI techniques in the current context is beyond the scope of this manuscript and will be the subject of future research. The results from the current study, obtained on three popular datasets in the field, indicate that the missing data mechanism has limited effect on fairness. Among the missing data handling methods, listwise deletion achieves the highest average fairness, while random forests yield the highest average fairness among the classification algorithms. Additionally, the interaction between the missing data handling technique and the classification algorithm is frequently significant.
\\
\\
To this end, the remaining sections are organised as follows: In Section 2 the relevant background material for fairness in ML and missing values is covered. In Section 3 a literature review of the limited work which combines the two fields of fairness and missing values is presented. In Section 4 the details of the experimental process are covered, while in Section 5 the results of the experiments are presented. Section 6 includes final conclusions and recommendations.

\section{BACKGROUND}
\subsection{Fairness in Machine Learning}
Fairness (or bias) of algorithms in ML is quantified by statistical definitions of so called \textbf{\textsl{fairness metrics}}, which are based on subsetting the outcomes of an algorithm in various ways to allow the difference in treatment of one group over another to be measured.
Broadly, there are two main categories of fairness metrics \citep{ven, zlio, verma, mehrabi, book, hardt}:
\begin{itemize}
\item Group fairness metrics: These aim to ensure non-discrimination across groups based on sensitive variables. 
\item Individual fairness metrics: These aim to give similar classification to similar individuals, where similarity is quantified by a distance metric.
\end{itemize}
Only group metrics will be considered, since this is a standard approach to evaluate fairness in ML. Details of individual fairness approaches can be found in \cite{dwork}.
\\
\\
Before providing the definitions of the group metrics considered in this study, some notation is introduced. Assume a binary classification scenario. Let $Y$ be the binary classification label taking values in \{0, 1\} where 1 is the positive outcome and 0 is the negative outcome. $S$ is the binary sensitive or protected attribute where 1 is the privileged group and 0 is the unprivileged group. $\hat{Y}$ are the predicted outcomes from a classifier, which are class predictions in \{0, 1\}. Assume that $S$ is fully observed, i.e. it contains no missing values.
\\
\\
The group metrics can be divided into two broad categories: those based on predicted outcome only and those based on both predicted and actual outcomes. 
\\
\\
A measure that is based on the predicted outcome only is \textbf{\textsl{demographic parity}}. This measure requires the probability for an individual to be assigned the favourable outcome to be equal across the privileged and unprivileged groups:
\begin{equation}
P( \hat{Y} = 1 \mid S = 1) = P( \hat{Y} = 1 \mid S = 0)
\end{equation}
Two instances of measures which are based on both predicted and actual outcomes are:
\begin{itemize} 
\item \textbf{\textsl{equality of opportunity}}, which requires equality of the true positive rate for both groups:
\begin{equation}
P( \hat{Y} = 1 \mid Y = 1, S = 1) = P( \hat{Y} = 1 \mid Y = 1, S = 0)
\end{equation}
\item \textbf{\textsl{predictive equality}}, which requires equality of the false positive rate for both groups:
\begin{equation}
P( \hat{Y} = 1 \mid Y = 0, S = 1) = P( \hat{Y} = 1 \mid Y = 0, S = 0)
\end{equation}
\end{itemize}
Focus is placed on these three metrics as they are commonly studied in the fairness literature \citep{hardt, verma, mehrabi}. Note that in practice, for the above metrics, the difference of the two terms in the constraints is taken to give a value for the difference in rates between the groups. Henceforth, the metrics demographic parity (\textit{dp}), predictive equality (\textit{pe}) or equality of opportunity (\textit{eo}) in the remainder of this article refer to this difference. Consider \textit{dp} as an example:
\begin{equation}
dp = P( \hat{Y} = 1 \mid S = 1) - P( \hat{Y} = 1 \mid S = 0) 
\end{equation}
 The smaller the absolute value of this difference, the smaller the discrimination and hence the higher the fairness. Take note that a negative value reflects `positive discrimination' in favour of the disadvantaged group. Other than the above mentioned three there also exist a variety of other group metrics \citep{book, zlio}, which is beyond the scope of this study. 

\subsection{Missing values} \label{miss-vals}
\textbf{\textsl{Missing data mechanisms}} (MDMs) \citep{rubin, enders, vanb} describe the relationship between the probability of missingness in a dataset and the variables in the dataset. To aid in the understanding of the MDMs, suppose a dataset contains the variables \textit{education level} and \textit{race}, and the probability of missingness in \textit{education level} is of interest. In this example, it is assumed that there is no missingness in the \textit{race} variable.
\\
\\
There are three MDMs: 
\begin{itemize}
    \item \textbf{\textsl{Missing completely at random}} (MCAR): The probability of being missing is unrelated to the data. In the given example, the observed values of \textit{education level} can be seen as a simple random sample of the complete variable. The missingness in \textit{education level} does not depend on either \textit{education level} or \textit{race}.
    \item \textbf{\textsl{Missing at random}} (MAR): The missingness probability only depends on the observed information. In the given example, suppose that the \textit{race} variable has two categories: white and non-white. The probability of missingness in \textit{education level} depends only on the respondent's \textit{race}.
    \item \textbf{\textsl{Missing not at random}} (MNAR): The probability of being missing depends on the missing information and can also depend on the observed information. In the given example, respondents with a lower \textit{education level} are less likely to give their level of education than those with a higher level of education.
\end{itemize}

The following missing data handling methods are applied in this study:

\begin{itemize}
\item \textbf{\textsl{Listwise deletion} (LD)}: This method simply removes the entire observation if missing values occur in any of the predictors. It is one of the most frequently employed techniques to deal with missing values as it is the default method of many statistical software packages. However, the disadvantage of a reduced sample size, followed by biased parameter estimation if the MDM is not MCAR, outweighs the advantage of it being a convenient and straightforward approach \citep{enders, kang}.
\item \textbf{\textsl{Single imputation}}:
The meaning of the term `single' in this context is that this imputation method generates a single replacement value per missing value. An advantage of imputation is the retention of those rows of data which contained missing values. Disadvantages are that most single imputation techniques produce biased parameter estimates and underestimate sampling error \citep{enders}. Some popular single imputation techniques are:
\begin{itemize}
\item \textbf{\textsl{Mode imputation}}:
The mode of a variable is calculated on the available cases of the variable and used to impute its missing values.

\item \textbf{\textsl{Regression imputation}}:
Here a set of regression equations is estimated to predict the incomplete variables from the complete variables.

\item \textbf{\textsl{\textit{k}-nearest neighbour imputation}}: In \textit{knn} imputation, we take the $k$ most similar observations to the missing observation (where similarity is measured by a distance metric); then for a continuous variable you replace the missing value by the mean of the $k$ nearest neighbours and the mode for a categorical variable \citep{knn}. 
\end{itemize}
\end{itemize}

\section{LITERATURE REVIEW} \label{lit-rev}
Relevant aspects of the limited available literature on the topic of the amalgamation of missing data and the fairness of ML algorithms are discussed to provide the basis of the study presented here.
\\
\\
\cite{ugduck} aimed to shed light on how missing data and fairness are related, whether the subsamples with missing data are more unfair or less unfair, and whether LD or imputation is the more appropriate procedure for application. Six benchmark datasets from the field of fairness in ML are used which contain missing values. Demographic parity (\textit{dp}) is the only fairness metric used. The experiments consistently show that the smallest absolute value of the metric is found on the subsets of instances containing only the missing value rows. The authors conclude that this result implies that the rows with missing values are fairer than the rest. This interpretation could however be somewhat misleading; the rows with missing data could potentially contain a higher proportion of individuals from both groups for whom the rate of favourable outcome is lower, as compared to the full set or the set without the missing rows. This could have resulted in a smaller metric value on the missing data rows. Furthermore, the trade-off between fairness and accuracy is explored. Overall, lower fairness (and higher accuracy) was found for imputation compared to LD. The impact of the MDM on fairness was not explicitly investigated. 
\\
\\
The study in \cite{zhanglong} investigates the impact of missing data imputation methods on fairness metrics. Seven imputation methods are investigated: MICE, missForest, \textit{knn}, two matrix completion methods (Soft-Impute and OptSpace), and two deep learning methods (Gain and Misgan). Most of these methods are state of the art and not widely known or used by practitioners. Results are based on only one of the fairness benchmark datasets. A novel notion is proposed for measuring the fairness of imputations, based on the imputation accuracy (which refers to the success of an imputation algorithm and is different from accuracy in the usual sense as a performance metric). The usefulness of this measure in regard to measuring fairness is unclear, as higher imputation accuracy does not necessarily mean higher fairness, especially if the dataset contains historic biases. The authors conclude that severe imputation unfairness exists among all the imputation methods, but the threshold of ‘severe’ has not been defined, and is usually context dependent. It is also observed that imputation unfairness tends to grow as the missingness proportion increases and imbalance in the size of sensitive classes leads to imputation unfairness. Importantly, they observe that prediction fairness is linked to the MDM, and there is a trade-off between accuracy and fairness in prediction.
\\
\\
A reweighting method for missing values in categorical predictors is considered by \cite{wangsingh}. The algorithm that is used as a starting point learns a probabilistic transformation to change feature value labels in the data to reduce discrimination, which is adapted with a reweighting scheme to account for missing values. Missingness is created according to the three MDMs in one of the categorical variables, but the specifications of the MDMs and variable selection are not provided. The empirical work is undertaken on two of the fairness benchmark datasets and one synthetic dataset. The authors analyse the impact of their reweighting scheme, different proportions of missingness and the three MDMs on accuracy and fairness. They conclude that the reweighting algorithm has very little impact on fairness for MCAR, whereas the biggest improvement in fairness is seen for MNAR. For MAR and MNAR the reweighting algorithm is able to mitigate the negative effects of the missing values on fairness with a small impact on accuracy.
\\
\\
Finally, the work in \cite{caton} considers fairness and performance metrics on three benchmark datasets, after performing a variety of imputation methods. The variables as well as the number of variables where missing data is introduced are both randomly determined; only MCAR missingness is created. The distributions of the metrics are explored and it is concluded that the choice of imputation strategy and classification algorithm significantly affects both performance and fairness. However, the authors combine the results from all three datasets before presenting them, which could have masked the effects of the different combinations of classification algorithm and imputation method when these differ across the datasets. The authors also perform a 3-way ANOVA to determine if there is any effect of three categorical factors on fairness and performance: imputation strategy, classification algorithm and percentage of missing values. It was found that the interaction effects between the imputation strategy and the classification algorithm are consistently significant. Hence it is concluded that different imputation strategies affect the performance and fairness metrics differently depending on which classification algorithm is used. 

\section{EXPERIMENTAL PROCESS} 
\subsection{Outline and aim of experiments} \label{exp}
The aim of this study is to investigate the effects of both the MDM (MCAR, MAR or MNAR) and the simpler missing data handling methods (LD and single imputation methods) on fairness, and furthermore to understand the relationship between fairness and accuracy in this context. Firstly, aspects of how this study departs from the existing literature (cf. Section \ref{lit-rev}) are described, followed by details on the experiments.
\\
\\
As mentioned in Section \ref{miss-vals}, this study investigates LD and the following single imputation methods: mode imputation, regression imputation and \textit{knn} imputation. The investigation in \cite{missingmethodsprop} shows that across disciplines most studies (81\%) used LD, 14\% of studies used a single imputation technique and MI was used in only 8\% of the studies. Also, the study in \cite{vim} highlights that in a usual statistical analysis, it is often sufficient to generate a single complete dataset which is then used in subsequent analyses. Hence LD and single imputation still have importance in the missing data field. Similar to \cite{ugduck}, the effect of LD and mode imputation on the fairness of ML models is studied, however, the context of the MDM is a core characteristic of this study.
\\
\\
Missingness is created in variables which are strongly related to the outcome variable; this is a common strategy when generating synthetic missing data according to the three MDMs \citep{santos}. To determine the relationship between predictors and the outcome variable, random forest variable importance is used to identify the importance of variables for the outcome variable. Take note, that `strongly related' variables refer to a relative strength of association between variables.  Missing data might not be evenly distributed between the privileged and unprivileged groups \citep{ugduck} since disadvantaged individuals may intentionally omit information if they believe that a complete answer might lead to a discriminatory or unfair action. To reflect this behaviour of non-disclosure of certain information, the variables in which missingness is created are chosen in a meaningful manner, which also allows the relationships between variables to be taken into account. This strategy is in contrast to previous studies \citep{zhanglong, caton}, where the variables in which missingness was created were chosen at random or without any context. In this study, the MAR missingness in a variable is created by dependence of missingness on the sensitive attribute; in particular we implement the assumption that observations from disadvantaged groups are more likely to contain missing values. 
\\
\\
In the research in \cite{zhanglong}, missingness was generated in numerical variables only. To represent a more realistic scenario, missingness is created in both numerical and categorical variables in this study, at most two variables at a time. In fairness related datasets there are often many categorical variables with missing values (as these are used to capture personal data) and this situation has not been studied in detail before \citep{ugduck, wangsingh}.
\\
\\
Three of the most widely used datasets in the field of fairness in ML are employed in this study: these datasets cover some pertinent real world fairness contexts: credit risk, the justice system and income level. All experiments are performed in {\tt R} \citep{R}. Missingness (i.e. amputation) is created according to the three MDMs, followed by either LD or single imputation (cf. Section \ref{miss-vals}). Four types of ML models are trained on this data: logistic regression, random forests, boosting and support vector machine (\textit{svm}), and the parameters of these models are tuned by cross-validation. Fairness metrics (\textit{dp}, \textit{pe} and \textit{eo}) and accuracy are calculated from the predictions on the test sets. The number of iterations varied between 20 to 100 (depending on the dataset), resulting in a distribution of fairness and accuracy values. Relevant details of the experiments are provided in the following sections. 

\subsection{Datasets}
Since the focus is on the impact of the missing data handling approach per MDM on fairness, all datasets should be complete at the onset of the experiment. To ensure this, rows containing missing values are removed prior to any analysis. All three datasets contain both numerical and categorical variables.
\\
\\
The \textit{German credit} dataset \citep{gc} classifies individuals described by a set of features as good or bad credit risks. The sensitive variables are \textit{sex} (male, female) and \textit{age} (young, old). The outcome variable is \textit{credit risk} (positive outcome: good, negative outcome: bad). It is a relatively small dataset containing no missing values, with 1000 observations and 20 predictor variables.
\\
\\
The \textit{Adult income} dataset \citep{adult} contains information about individuals from the 1994 United States census. This publicly available dataset is published as a separated training and test set, which is combined into one dataset to allow for a new assignment of the subsets. The sensitive variables are \textit{sex} (male, female) and \textit{race} (white, non-white). The outcome variable is whether an individual makes more or less than \$50,000 in yearly income (positive outcome: $>$ \$50,000). It is a relatively large dataset with 13 predictors and 45222 observations remaining after the removal of 3620 missing value rows. 
\\
\\
The \textit{COMPAS} dataset \citep{compas} refers to data collected about the use of the COMPAS risk assessment tool in Broward County, Florida. The dataset contains 7214 observations and 52 predictors; only 11 predictors containing no missing values are used for experimentation, as was done in \cite{2022survey}. The sensitive variables are \textit{race} (white, non-white) and \textit{sex} (male, female). As in \cite{ugduck} the females are regarded to be the privileged group. The outcome variable is \textit{two year recid} indicating whether offenders were rearrested within two years after the first arrest, with not being rearrested as the positive outcome.

\subsection{Amputation}
The task of data amputation in a complete dataset is undertaken with the {\tt ampute} function \citep{ampute} in the {\tt R} package {\tt mice} \citep{mice}. The {\tt ampute} function allows the user to generate missing values in multiple variables, with different missingness mechanisms. The merits of using the multivariate {\tt ampute} function rather than the widespread practice of generating missingness in one variable at a time are detailed in \cite{ampute} along with the pitfalls of the univariate amputation approach. The study in \cite{caton} tested a range of missing percentages between 1 and 10\% and found that in this range the percentage of missing values does not significantly affect the fairness and accuracy, hence it was decided to choose a more extreme missingness percentage, in particular 50\%. The default percentage of missing values that is generated in the {\tt ampute} function is also 50\% \citep{mice}. Choosing such a relatively high value for the proportion of missingness may aid in detecting the effects of amputation more clearly.

\subsection{Imputation}
The missing data methods that will be implemented are LD and three popular single imputation techniques: mode imputation (since missingness will be created in categorical variables as well as numerical variables), regression imputation and \textit{knn} imputation. Both regression imputation and \textit{knn} imputation will use the sensitive variable when creating the regression model or calculating distance in \textit{knn}; since the imputation model does not use the sensitive variable to differentiate between the groups in terms of predictions, it is a reasonable approach to take. The outcome variable will not be used in the imputation models. 

\subsection{Training and test sets} \label{ts}
Each of the amputed datasets is randomly split into an amputed training set and an amputed test set. Using the same indices, the original complete dataset (which has no missing values) is split into a complete training set and complete test set. The test set is a third of the dataset. The amputed training sets are processed according to an imputation method or LD. We use the complete test sets to collect the results.

\subsection{Classification models}
Model selection is performed on the training set using 5-fold cross-validation, and in line with standard practice parameter tuning is done with the aim of maximising prediction accuracy. Regarding the range of classification models, four popular models are implemented. Logistic regression gives a linear separation boundary but is not very flexible; random forests are flexible and a good fit for non-linear separation; boosting (trees) again give a very flexible model; \textit{svm} is robust to outliers and the radial kernel method gives a flexible model. To align with standard practice in industry (often in compliance with legal requirements), the classification model will not use the sensitive variable as a predictor \citep{inc_s}. The complete test sets are used to determine the fairness and accuracy of the predictions after amputation, missing data handling and classification.

\section{RESULTS}
In the previous section, the experimental process was described for the investigation of the effects of both the MDM and simple missing data handling methods on the fairness (and accuracy) of ML models. In this section, we make an empirical comparison of fairness metrics as impacted by the factors under investigation. We employ analysis of variance (ANOVA) and boxplots in order to compare distributions of fairness metrics across MDMs, imputation strategies and classification models. To aid the decision of whether to apply parametric ANOVA or a non-parametric version, suitable tests are conducted on the fairness distributions within the ANOVA groups to determine whether the assumptions of normality and homoscedasticity across groups are satisfied. For cases in which there was evidence of the violation of homoscedasticity, the non-parametric Welch-ANOVA was applied. Boxplots are utilised to demonstrate which combinations or main effects lead to the highest or lowest fairness and accuracy.

\begin{table*}[hbt!]
\centering
\setlength{\tabcolsep}{6pt}   
\small
\begin{threeparttable}
\caption{\textit{p}-values of main and interaction effects of the MDM, imputation method (`imp') and classification model (`mod') from ANOVA/Welch-ANOVA analysis across the datasets, sensitive variables (`S') and metrics (\textit{dp}, \textit{pe}, \textit{eo} and accuracy (acc)).}
\label{tab:results}

\begin{tabular}{
l l l
p{1.4cm}
p{1.45cm}
p{1.45cm}
S[table-format=1.4]
S[table-format=1.4]
S[table-format=1.4]
S[table-format=1.4]
}
\toprule
Dataset & S & Metric & {MDM} & {imp} & {mod} & {mdm*imp} & {mdm*mod} & {imp*mod} & {mdm*imp*mod} \\
\midrule

\multirow{8}{*}{German}
& \multirow{4}{*}{Sex}
& dp  & 0.8932 & 0.0274* & 0.5050 & 0.8912 & 0.8568 & 0.9570 & 0.9782 \\
&     & pe  & 0.8682 & 0.1195  & 0.1749 & 0.7701 & 0.8153 & 0.9106 & 0.9666 \\
&     & eo\dag  & 0.3171 & 0.0454* & 0.0000*** & 0.7642 & 0.4676 & 0.0857 & 0.9948 \\
&     & acc & 0.2373 & 0.0000*** & 0.0000*** & 0.7468 & 0.6561 & 0.7308 & 0.9997 \\
\cmidrule(lr){2-10}
& \multirow{4}{*}{Age}
& dp\dag  & 0.0329 & 0.0000*** & 0.0000*** & 0.0238* & 0.4550 & 0.0006*** & 0.9969 \\
&     & pe  & 0.7658 & 0.0012** & 0.0003*** & 0.8310 & 0.9759 & 0.4211 & 0.9992 \\
&     & eo\dag  & 0.0677 & 0.0001*** & 0.0000*** & 0.0914 & 0.5562 & 0.0305* & 0.9764 \\
&     & acc & 0.6383 & 0.0000*** & 0.0067** & 0.5755 & 0.6694 & 0.6783 & 0.9991 \\

\midrule
\multirow{8}{*}{Adult}
& \multirow{4}{*}{Sex}
& dp\dag  & 0.0000*** & 0.0000*** & 0.0000*** & 0.0040** & 0.5566 & 0.0131* & 0.4442 \\
&     & pe\dag  & 0.0001*** & 0.0000*** & 0.0000*** & 0.0002*** & 0.6107 & 0.0049** & 0.8050 \\
&     & eo  & 0.5481 & 0.0063** & 0.0000*** & 0.4068 & 0.8913 & 0.0000*** & 0.9590 \\
&     & acc & 0.8476 & 0.2847 & 0.0000*** & 0.9976 & 0.3311 & 0.0000*** & 0.9694 \\
\cmidrule(lr){2-10}
& \multirow{4}{*}{Race}
& dp  & 0.6678 & 0.3207 & 0.0016*** & 0.9161 & 0.9513 & 0.4172 & 0.8063 \\
&     & pe\dag  & 0.0293* & 0.0000*** & 0.0000*** & 0.0719 & 0.8896 & 0.0581 & 0.9871 \\
&     & eo  & 0.9338 & 0.9862 & 0.0000*** & 0.9998 & 0.9599 & 0.4850 & 1.0000 \\
&     & acc & 0.9712 & 0.7509 & 0.0000*** & 0.9809 & 0.7199 & 0.0000*** & 0.7768 \\

\midrule
\multirow{8}{*}{COMPAS}
& \multirow{4}{*}{Sex}
& dp  & 0.6904 & 0.0011** & 0.0000*** & 0.9964 & 0.9984 & 0.0851 & 0.9947 \\
&     & pe  & 0.8062 & 0.1421 & 0.0030** & 0.9473 & 0.9906 & 0.5022 & 0.9953 \\
&     & eo  & 0.4939 & 0.0020** & 0.0000*** & 0.9297 & 0.9632 & 0.0509 & 0.9987 \\
&     & acc & 0.6672 & 0.1703 & 0.0000*** & 0.9826 & 0.9319 & 0.0056** & 0.9977 \\
\cmidrule(lr){2-10}
& \multirow{4}{*}{Race}
& dp  & 0.7125 & 0.0450* & 0.0093** & 0.9489 & 0.9980 & 0.9679 & 0.9999 \\
&     & pe  & 0.7962 & 0.0363* & 0.0003*** & 0.9511 & 0.9918 & 0.7515 & 1.0000 \\
&     & eo  & 0.7874 & 0.1550 & 0.0134* & 0.8632 & 0.9994 & 0.9946 & 1.0000 \\
&     & acc & 0.9509 & 0.0224* & 0.0000*** & 0.9376 & 0.9585 & 0.2180 & 0.9944 \\

\bottomrule
\end{tabular}

\begin{tablenotes}
\footnotesize
\item $^{*} p < 0.05$, $^{**} p < 0.01$, $^{***} p < 0.001$. \dag \ indicates cases where Welch-ANOVA analysis was used
\end{tablenotes}
\end{threeparttable}
\end{table*}

The results of the ANOVA analyses over the three datasets are summarised in Table \ref{tab:results}. The higher-order interactions (i.e. three-way interactions) are interpreted first, followed by the second-order interactions and main effects in the absence of significant interaction.  Take note of the additional notation introduced for this section: \textit{lr} for logistic regression, \textit{rf} for random forests, \textit{b} for boosting and \textit{reg} for regression imputation.

\subsection{Three-way interaction}
From Table \ref{tab:results} we see that there is no significant interaction between the MDM, imputation method and classification model across all datasets, metrics and sensitive variables. Therefore, the manner in which two of the factors interact does not depend on the third factor.

\subsection{Two-way interactions}
From Table \ref{tab:results} it is observed that there is some evidence of two-way interactions between the factors. In particular: the interaction between imputation method and classification model (imp*mod), and the interaction between the MDM and the imputation method (mdm*imp). No significant interaction is observed between the MDM and classification model (mdm*mod). 

\subsubsection{Imputation method and classification model}
\begin{figure}[h!]
       \centering       \includegraphics[width=1\textwidth]{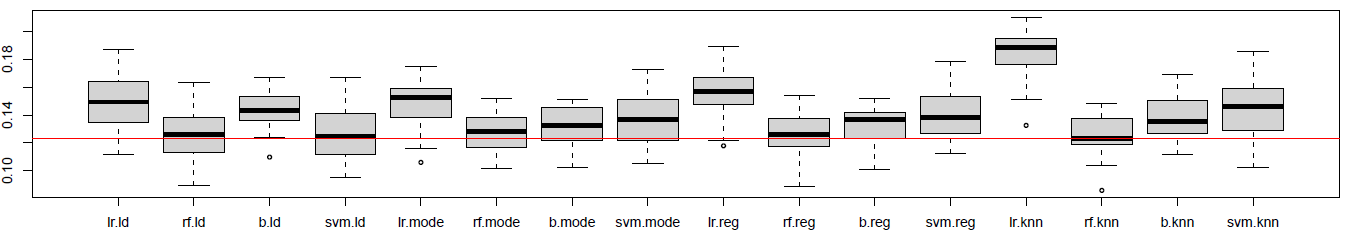}
       \caption{Adult dataset, sensitive variable sex, MAR, equality of opportunity (\textit{eo}) distributions}
		\label{fig1}
   \end{figure}

\begin{figure}[h!]
       \centering       \includegraphics[width=1\textwidth]{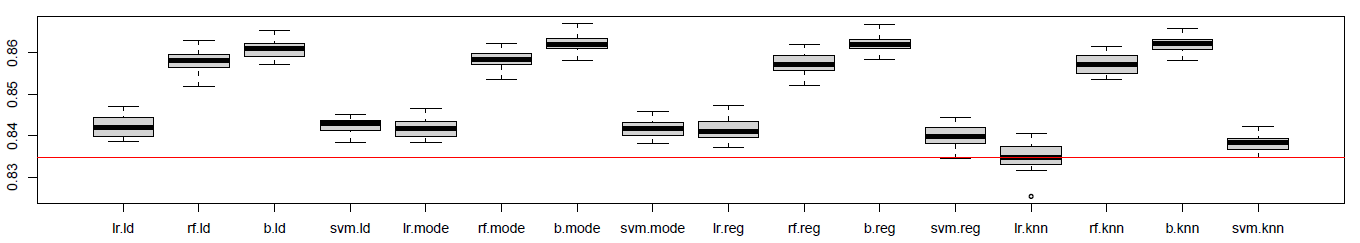}
       \caption{Adult dataset, MAR, accuracy distributions}
		\label{fig2}
   \end{figure}

\begin{figure}[h!]
       \centering       \includegraphics[width=1\textwidth]{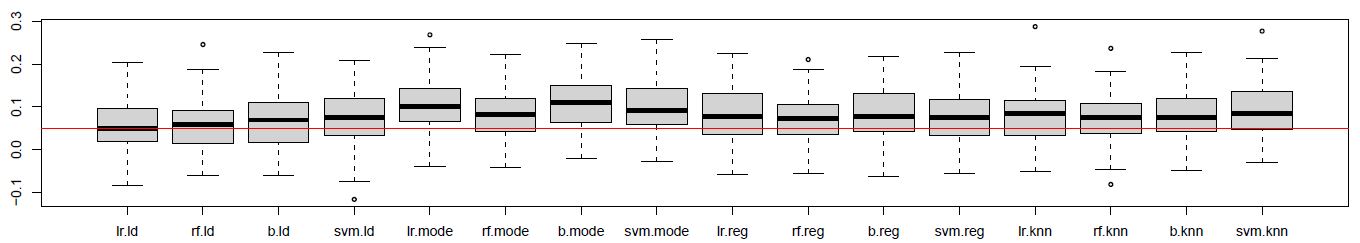}
       \caption{German dataset, sensitive variable age, MAR, demographic parity (\textit{dp}) distributions}
		\label{fig3}
   \end{figure}

\begin{figure}[h!]
       \centering       \includegraphics[width=1\textwidth]{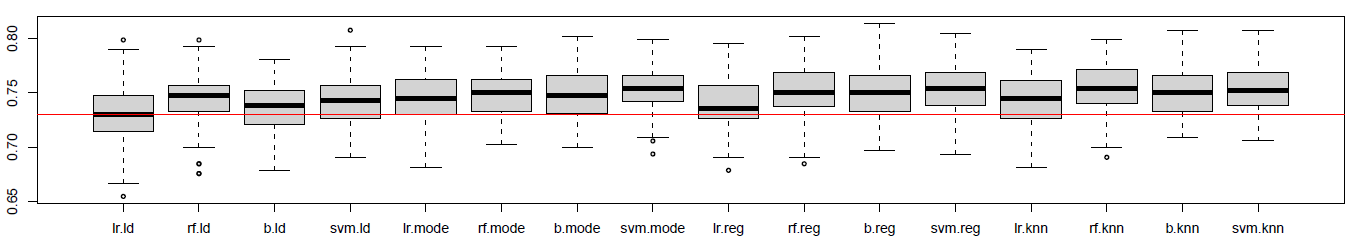}
       \caption{German dataset, MAR, accuracy distributions}
		\label{fig4}
   \end{figure}

It is interesting to note that the significant interaction between the imputation method and classification model occurred mostly for the Adult dataset (for sensitive variable \textit{sex}) across all metrics, and for the German dataset (for sensitive variable \textit{age}) for \textit{dp} and to a lesser extent, \textit{eo}. To illustrate this, figures showing all possible combinations of the missing data handling techniques and the classification models, for a selection of MDMs, are presented. 16 boxplots are presented in a figure, with the x-axis label order as follows: lr.ld, rf.ld, b.ld, svm.ld, lr.mode, rf.mode, b.mode, svm.mode, lr.reg, rf.reg, b.reg, svm.reg, lr.knn, rf.knn, b.knn, svm.knn. This enables the comparison of fairness across classification models, after the application of each missing data handling technique.
In order to aid interpretation, a red horizontal line is displayed intersecting the smallest median discrimination (highest median fairness), as obtaining high fairness is often of interest in this context. Similarly, for the plots displaying accuracy, the red horizontal line is positioned at the smallest median accuracy, as the tradeoff between fairness and
accuracy for a particular combination is often of interest. 
\\
\\
Aggregating over the significant results in Table \ref{tab:results}, the following overall patterns regarding the interaction between imputation method and classification model are observed:
\begin{itemize}
\item Across imputation techniques, in the majority of cases the \textit{rf} model gives the highest fairness on average, with the combination of LD and \textit{rf} often giving the highest fairness. This is illustrated in Figure \ref{fig1} for the Adult dataset for \textit{eo} where \textit{rf} consistently gives the highest fairness on average for all imputation techniques. Figure \ref{fig3} illustrates this for \textit{dp} for the German dataset but the distinctions are not as clear.
\item Across imputation techniques, in the majority of cases the \textit{lr} model gives the lowest fairness on average. Figure \ref{fig1} illustrates this for the Adult dataset for \textit{eo}, where the combination of \textit{knn} and \textit{lr} result in the lowest fairness on average. In Figure \ref{fig3}, for the German dataset for \textit{dp} we see that the combination of mode and \textit{lr} results in one of the lowest fairness on average.
\item Regarding the trade-off between fairness and accuracy, the effects differ between the Adult and German dataset. For the Adult dataset, rather than a trade-off, we see the combination which results in the lowest fairness (\textit{knn} and \textit{lr}) also leads to the lowest accuracy, and the \textit{rf} models have on average both higher fairness and accuracy. This can be seen in Figures \ref{fig1} and \ref{fig2}. For the German dataset, there appears to be a trade-off such that higher fairness combinations are associated with lower accuracy. This is illustrated in Figures \ref{fig3} and \ref{fig4}.
\end{itemize}

\subsubsection{MDM and imputation method}
\begin{figure}[h!]
       \centering       \includegraphics[width=1\textwidth]{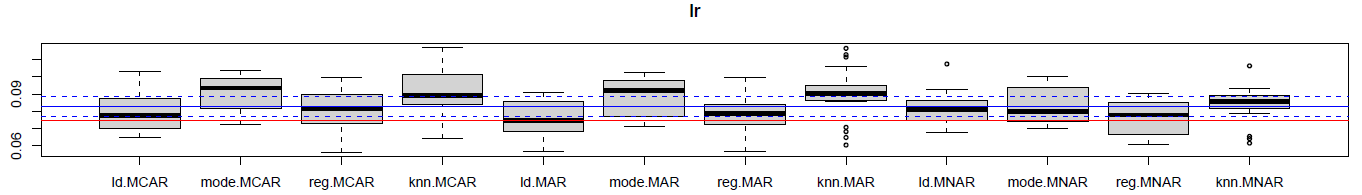}
       \caption{Adult dataset, sensitive variable sex, \textit{lr} model, predictive equality (\textit{pe}) distributions}
		\label{fig5}
   \end{figure}

\begin{figure}[h!]
       \centering       \includegraphics[width=1\textwidth]{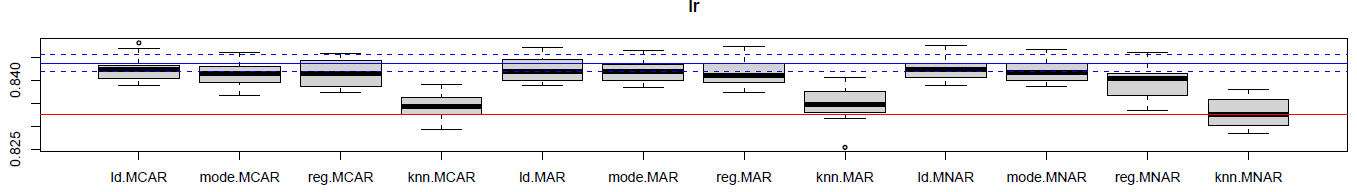}
       \caption{Adult dataset, \textit{lr} model, accuracy distributions}
		\label{fig6}
   \end{figure}

\begin{figure}[h!]
       \centering       \includegraphics[width=1\textwidth]{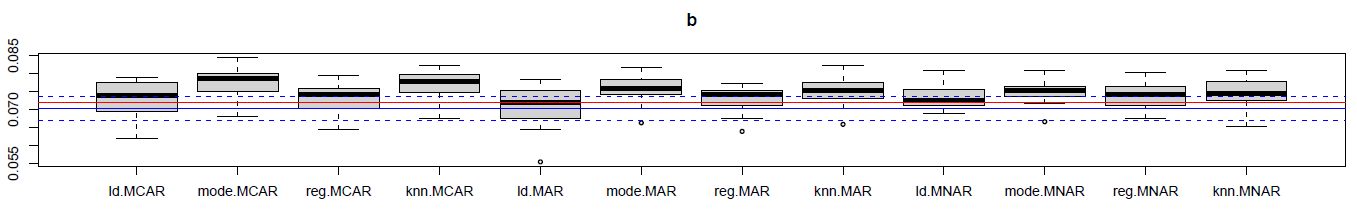}
       \caption{Adult dataset, sensitive variable sex, \textit{b} model, predictive equality (\textit{pe}) distributions}
		\label{fig7}
   \end{figure}

\begin{figure}[h!]
       \centering       \includegraphics[width=1\textwidth]{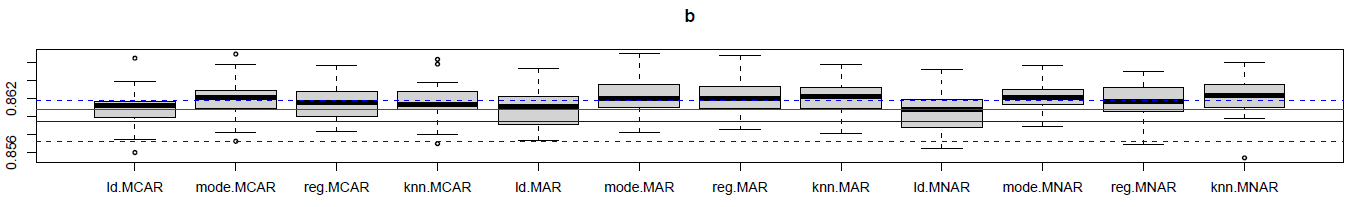}
       \caption{Adult dataset, \textit{b} model, accuracy distributions}
		\label{fig8}
   \end{figure}

There is limited evidence for the interaction effect between the MDM and the imputation method; the significant results are observed for the Adult dataset, sensitive variable sex and metrics \textit{dp} and  \textit{pe}, and the German dataset, sensitive variable age and metric \textit{dp}. Again, figures are utilised to show all possible combinations of the MDMs and the imputation techniques, for the four classification models. In this section, figures are only presented for the Adult dataset as the significant differences are easier to observe by eye for this dataset, as opposed to the German dataset. 12 boxplots are presented in a figure, with the
x-axis label order as follows: ld.MCAR, mode.MCAR, reg.MCAR, knn.MCAR, ld.MAR, mode.MAR, reg.MAR, knn.MAR, ld.MNAR, mode.MNAR, reg.MNAR and knn.MNAR. Therefore, the fairness across the imputation techniques, within each MDM, can be compared. As for the previous section, a red horizontal line illustrates the smallest median discrimination and similarly the smallest median accuracy. In addition, \textit{baseline} metric values per classification model are presented in blue: the mean and standard deviation of the fairness (and accuracy) distributions, obtained by fitting classification models on the complete training and complete test sets, such that no amputation or imputation is performed on these datasets. This provides insight to understand how much the discrimination increases through the missing data pipeline. 
\\
\\
The following overall patterns in the average fairness and accuracy are observed when considering the interaction between the MDM and the imputation method: 
\begin{itemize}
    \item Across the MDMs, LD gives the highest fairness on average, with the combination of MAR and LD resulting in the highest fairness overall, as illustrated in Figures \ref{fig5} and \ref{fig7}. 
    \item Mode or \textit{knn} give the lowest fairness on average across the MDMs, with the combination of MCAR and mode resulting in the lowest fairness overall, as illustrated in Figures \ref{fig5} and \ref{fig7}. Interestingly, the distributions for mode and \textit{knn} tend to exceed 1 standard deviation of the average baseline level for the classification model.
    \item There does not appear to be a trade-off with accuracy in general (see Figures \ref{fig6} and \ref{fig8}); lower fairness tends to correlate with lower accuracy, etc. For the \textit{lr} model, the \textit{knn} accuracy distributions deviate noticeably from the baseline. 
\end{itemize}

\subsection{Main effects}
After accounting for the significant interactions in Table \ref{tab:results}, it can be seen that there is no significant main effect of the MDM on the fairness and accuracy distributions. Therefore, only the main effects of the imputation method and classification model are investigated. Across datasets and metrics, in the majority of cases the classification model has a significant main effect, followed by the main effect of the imputation model. The effect of the imputation model is investigated first, since in the experimentation pipeline handling the missing data comes before fitting a classifier. Whereas the two-way interaction effects were significant for a limited subset of the datasets, sensitive variables and metrics, the main effects in general apply across these three items.

\subsubsection{Imputation method}
\begin{figure}[h!]
    \centering
    \begin{subfigure}[t]{0.26\textwidth}
        \centering        \includegraphics[width=\linewidth]{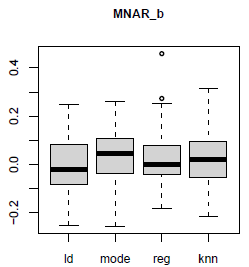}
        \caption{German dataset, sensitive variable age, predictive equallity (\textit{pe}) distributions}
        \label{fig9_1}
    \end{subfigure}
        \hspace{1.3cm}
    \begin{subfigure}[t]{0.248\textwidth}
        \centering        \includegraphics[width=\linewidth]{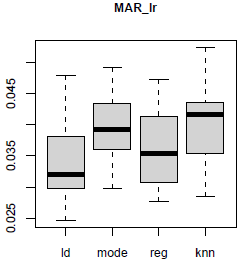}
        \caption{Adult dataset, sensitive variable race, predictive equality (\textit{pe}) distributions}
        \label{fig9_2}
    \end{subfigure}
    \hspace{1.3cm}
    \begin{subfigure}[t]{0.249\textwidth}
        \centering        \includegraphics[width=\linewidth]{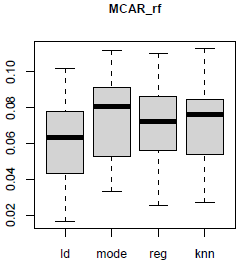}
        \caption{COMPAS dataset, sensitive variable sex, equality of opportunity (\textit{eo}) distributions}
        \label{fig9_3}
    \end{subfigure}
    \caption{Main effect of missing data handling technique}
    \label{fig9}
\end{figure}

Regarding the main effect of the missing data handling method, aggregating across the significant results in Table \ref{tab:results} the following patterns are observed:
\begin{itemize}
    \item The highest fairness on average is observed most often for LD. For illustration, some relevant examples from the three datasets are provided in Figure \ref{fig9}.
    \item The lowest fairness on average is observed for mode imputation or \textit{knn} imputation (see Figure \ref{fig9}).
    \item Regarding the trade-off between fairness and accuracy, this is again dataset dependent as seen in the previous sections, so we do not provide an illustration; as before there appears to be a trade-off in the German dataset, but no clear trade-off in the Adult and COMPAS datasets. 
\end{itemize}
These observations are consistent with the findings from the two-way interactions, where LD results in the highest fairness on average and mode or \textit{knn} imputation results in the lowest fairness on average.

\subsubsection{Classification model}
\begin{figure}[h!]
    \centering
    \begin{subfigure}[t]{0.26\textwidth}
        \centering        \includegraphics[width=\linewidth]{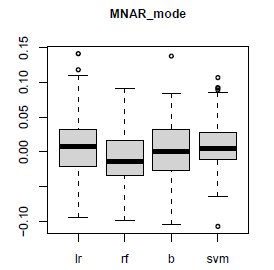}
        \caption{German dataset, sensitive variable sex, equality of opportunity (\textit{eo}) distributions}
        \label{fig10_1}
    \end{subfigure}
        \hspace{1.3cm}
    \begin{subfigure}[t]{0.25\textwidth}
        \centering        \includegraphics[width=\linewidth]{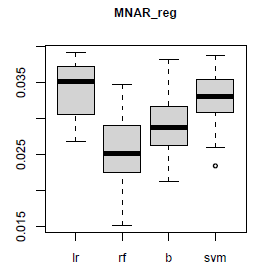}
        \caption{Adult dataset, sensitive variable race, predictive equality (\textit{pe}) distributions}
        \label{fig10_2}
    \end{subfigure}
    \hspace{1.3cm}
    \begin{subfigure}[t]{0.25\textwidth}
        \centering        \includegraphics[width=\linewidth]{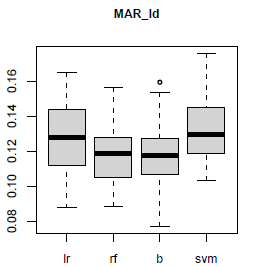}
        \caption{COMPAS dataset, sensitive variable sex, equality of opportunity (\textit{eo}) distributions}
        \label{fig10_3}
    \end{subfigure}
    \caption{Main effect of classification model}
    \label{fig10}
\end{figure}

\begin{figure}[h!]
    \centering
    \begin{subfigure}[t]{0.265\textwidth}
        \centering        \includegraphics[width=\linewidth]{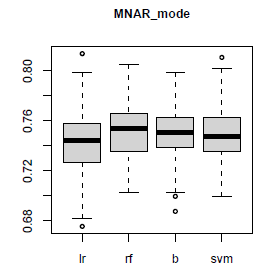}
        \caption{German dataset}
        \label{fig11_1}
    \end{subfigure}
        \hspace{1cm}
    \begin{subfigure}[t]{0.27\textwidth}
        \centering        \includegraphics[width=\linewidth]{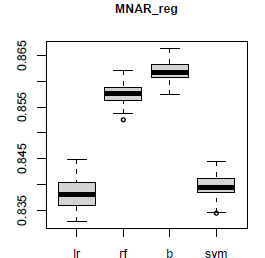}
        \caption{Adult dataset}
        \label{fig11_2}
    \end{subfigure}
    \hspace{1cm}
    \begin{subfigure}[t]{0.25\textwidth}
        \centering        \includegraphics[width=\linewidth]{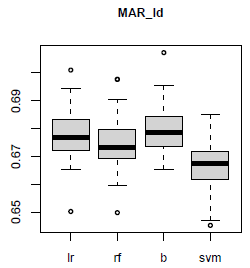}
        \caption{COMPAS dataset}
        \label{fig11_3}
    \end{subfigure}
    \caption{Accuracy distributions for main effect of classification model}
    \label{fig11}
\end{figure}

We now investigate the main effect of the classification model. Aggregating across the significant results in Table \ref{tab:results} the following patterns are observed:
\begin{itemize}
    \item The highest fairness on average is observed for the \textit{rf} model, followed by the \textit{b} model. This is illustrated in Figure \ref{fig10} for some relevant examples across the MDMs and missing data handling techniques.
    \item The lowest fairness on average is observed for the \textit{lr} model, followed by the \textit{svm} model (see Figure \ref{fig10}).
    \item Interestingly, there is no clear trade-off observed between fairness and accuracy across the datasets when the main effect of the classification model is investigated. Overall, the highest average accuracy is obtained for the \textit{rf} or \textit{b} model and the lowest average accuracy is obtained for the \textit{lr} or \textit{svm} model. This is illustrated in Figure \ref{fig11} for the corresponding examples from Figure \ref{fig10}.
\end{itemize}

\subsection{Main findings}
The main findings from this study can be summarised as follows:
\begin{itemize}
    \item There is insufficient evidence for a three-way interaction between MDM, missing data handling method and classification model.
    \item In the very limited number of cases where there is a significant two-way interaction between MDM and missing data handling method, the highest fairness on average is found for MAR and LD, and the lowest fairness on average is obtained for MCAR and mode imputation.
    \item In the small number of cases where there is a significant two-way interaction between missing data handling method and classification model, it is observed that the combination of LD and the \textit{rf} model give the highest fairness on average, and the lowest fairness on average is obtained for the combination of either mode or \textit{knn} imputation with the \textit{lr} model.
    \item For the sizeable number of cases where the main effect of the imputation model is significant, LD gives the highest fairness on average, and mode or \textit{knn} give the lowest fairness on average.
    \item The main effect of the classification model is significant for the majority of cases; the \textit{rf} model gives the highest fairness on average, and the \textit{lr} model gives the lowest fairness on average.
    \item Regarding the trade-off between fairness and accuracy, this was observed for the German dataset but not for the other two. In general, the highest accuracy on average was often obtained for the \textit{rf} model and the lowest accuracy on average was often obtained for \textit{knn} imputation with the \textit{lr} model.
\end{itemize}

From the findings, it can be concluded that the MDM does not significantly affect the fairness and accuracy. It would appear that the downstream processes in the experimental pipeline `correct' for the effects of the MDM. The study by \cite{dance} may also shed some light in this context; it was shown that when dataset correlations are either low or very substantial, distinct MDMs can produce similar statistical inferences. 
\\
\\
Across imputation techniques, LD gives the highest fairness on average, with mode and \textit{knn} imputation resulting in the lowest fairness on average. It is interesting to note that the simpler method of row deletion leads to higher fairness than the more complex task of imputing the missing values. This finding aligns with the outcome of the study by \cite{ugduck}. Across classification models, the \textit{rf} model gives the highest fairnes on average and the \textit{lr} model gives the lowest fairness on average. In cases where the interaction of the classification model and missing data handling method is significant, the highest/lowest fairness is obtained for combinations of the main effects which result in the highest/lowest fairness; it is found that LD and \textit{rf} give the highest fairness on average, and \textit{knn} and \textit{lr} give the lowest fairness on average.
\\
\\
Regarding the trade-off between fairness and accuracy, this is dataset dependent; with the effect only apparent for the German dataset. For the Adult and COMPAS datasets, there is often a positive correlation between the fairness and accuracy so that when one is relatively large then so is the other; for example for the Adult dataset it is observed that \textit{knn} and \textit{lr} give the lowest fairness and lowest accuracy. It is also seen from the boxplots that the distinctions in the distributions between the categories of main or interaction factors are not as contrasted for the German dataset as for the other two datasets. 
\\
\\
Another interesting outcome is that the findings apply across the three metrics \textit{dp}, \textit{pe} and \textit{eo}. It can be shown that if the base rates of the privileged and unprivileged groups are equal, then \textit{dp} is a weighted average of \textit{pe} and \textit{eo}. Refer to Appendix \ref{relation} for further details.

\section{CONCLUSIONS, RECOMMENDATIONS AND FUTURE WORK}
The aim of this research was to understand the effects of missing data and missing data handling techniques on the fairness of ML algorithms. Furthermore, to consider how missing data and fairness are related. Given the increasing impact of such algorithms in daily life, understanding their fairness is vital.
\\
\\
A finding of this study is that the MDM does not seem to impact the fairness and accuracy. This may provide some relief for data practitioners encountering missing values in real-world datasets when it is usually not possible to determine the MDM for missing values. This study showed that careful choices of missing data handling method and classification model can limit the impact of the missing data. 
\\
\\
The main effect of the classification model proved to be significant overall, with the random forests model outperforming other classifiers, resulting in the highest average fairness and accuracy. This gives some indication that tree-based flexible models may be suitable to use when the aim is to improve fairness, with little cost to the accuracy. To the contrary, models like logistic regression which are inflexible should be avoided.
\\
\\
The main effect of the missing data handling technique was significant in a large number of cases, and LD achieved the highest average fairness overall, as compared to imputation, with minimum deleterious effect on the accuracy. LD is the simplest and least computationally expensive approach to deal with missing values compared to the other methods considered in this study. It is also often the default option to deal with missing values in statistical software. Given its widespread use, it is reassuring to observe its impact on fairness. The interaction of the missing data handling technique and classification model was also often significant and it was found overall that main effects which gave rise to high fairness also led to interaction effects with high fairness. 
\\
\\
Given that this study used real-world datasets in a simulation study, an extension of the work, inspired by \cite{dance}, could be to further investigate the impact of the MDM by using simulated datasets to control correlations between variables. A different choice of fairness metrics could also be considered as another extension to the current study, where the metrics are not so closely related to each other. As a final direction of research, it is proposed to conduct an analogous study to understand the impact of more advanced imputation methods such as MI on fairness to establish whether there are improvements in fairness when applying MI compared to the simple approaches in the current study. Careful attention should be paid to the combination of the fairness metric estimates in the analysis and pooling phases of MI.

\section{Declarations}
\subsection{Funding}
No funding was received to assist with the preparation of this manuscript.
\subsection{Competing interests}
The authors have no relevant financial or non-financial interests to disclose. The authors have no competing interests to declare that are relevant to the content of this article.

\appendix
\section{Relation between \textit{dp}, \textit{pe} and \textit{eo}} \label{relation}
We have that 
$$dp = P( \hat{Y} = 1 \mid S = 1) - P( \hat{Y} = 1 \mid S = 0) $$

By the law of total probability we can expand this out as
$$= P( \hat{Y} = 1 \mid Y = 1, S = 1)P(Y = 1 \mid S = 1) + 
P( \hat{Y} = 1 \mid Y = 0, S = 1)P(Y = 0 \mid S = 1) - $$
\begin{equation}
[P( \hat{Y} = 1 \mid Y = 1, S = 0)P(Y = 1 \mid S = 0) + 
P( \hat{Y} = 1 \mid Y = 0, S = 0)P(Y = 0 \mid S = 0)]
\end{equation}
$$= P( \hat{Y} = 1 \mid Y = 1, S = 1)P(Y = 1 \mid S = 1) - 
P( \hat{Y} = 1 \mid Y = 1, S = 0)P(Y = 1 \mid S = 0) + $$
\begin{equation}
[P( \hat{Y} = 1 \mid Y = 0, S = 1)P(Y = 0 \mid S = 1) -
P( \hat{Y} = 1 \mid Y = 0, S = 0)P(Y = 0 \mid S = 0)]
\end{equation}
Now, if we have that
\begin{equation}
P(Y = 1 \mid S = 1) = P(Y = 1 \mid S = 0) = P(Y = 1)
\end{equation}
which implies that
\begin{equation}
P(Y = 0 \mid S = 1) = P(Y = 0 \mid S = 0) = P(Y = 0),
\end{equation}
then recalling the definition of \textit{pe} and \textit{eo} we have that
\begin{equation}
dp = P(Y = 1)eo + P(Y = 0)pe
\end{equation}

Therefore, if the base rates of the privileged and unprivileged groups are equal, \textit{dp} is a weighted average of \textit{eo} and \textit{pe}.

\section{Some details of the amputation}
To create missing data in the three datasets according to the MDMs, the {\tt ampute} function \citep{ampute} in the {\tt R} package {\tt mice} \citep{mice} is used. For comparability, the variables per MDM in which missing values are created are the same for each dataset. MAR missingness is always created with dependency only on the sensitive variable. For the three datasets, the following are the variables in which missing values are created:
\begin{itemize}
    \item German dataset: checking account status, credit history
    \item Adult dataset: capital gain
    \item COMPAS dataset: age, score text
\end{itemize}

\bibliography{references}  
\end{document}